\documentclass{article}

\usepackage[preprint,nonatbib]{neurips_2020}

\usepackage[utf8]{inputenc} 
\usepackage[T1]{fontenc}    
\usepackage{hyperref}       
\usepackage{url}            
\usepackage{booktabs}       
\usepackage{amsfonts}       
\usepackage{nicefrac}       
\usepackage{microtype}      
\usepackage[numbers,sort&compress]{natbib}      

\usepackage{mathtools}
\usepackage{tabularx}
\usepackage{amsmath}
\usepackage{amssymb}
\usepackage[noend]{algpseudocode}
\usepackage{algorithm,algorithmicx}
\usepackage{graphicx}
\usepackage{color}
\usepackage{subfig}
\usepackage{booktabs}
\usepackage{caption}
\usepackage{array}
\usepackage[final]{pdfpages}
\usepackage{siunitx}
\usepackage{float}
\usepackage{wrapfig}
\usepackage{dblfloatfix}
\usepackage[nameinlink]{cleveref}
\usepackage{pdfpages}

\title{MetaSDF: Meta-learning Signed Distance Functions}

\author{%
	Vincent Sitzmann\thanks{These authors contributed equally to this work.} \\
	Stanford University\\
	\texttt{sitzmann@cs.stanford.edu} \\
	\And
	Eric R. Chan\footnotemark[1]\\
	Stanford University\\
	\texttt{erchan@stanford.edu}
	\And
	Richard Tucker\\
	Google Research\\
	\texttt{richardt@google.com}
	\And
	Noah Snavely\\
	Google Research\\
	\texttt{snavely@google.com}
	\And
	Gordon Wetzstein\\
	Stanford University\\
	\texttt{gordon.wetzstein@stanford.edu}
}

\begin{document}

\maketitle

\vbox{%
	\vskip -0.22in
	\hskip -0.15in
	\hsize\textwidth
	\linewidth\hsize
	\centering
	\tt\normalsize\href{https://vsitzmann.github.io/metasdf/}{vsitzmann.github.io/metasdf/}
	\vskip 0.2in
}

\begin{abstract}
Neural implicit shape representations are an emerging paradigm that offers many potential benefits over conventional discrete representations, including memory efficiency at a high spatial resolution.
Generalizing across shapes with such neural implicit representations amounts to learning priors over the respective function space and enables geometry reconstruction from partial or noisy observations.
Existing generalization methods rely on conditioning a neural network on a low-dimensional latent code that is either regressed by an encoder or jointly optimized in the auto-decoder framework.
Here, we formalize learning of a shape space as a meta-learning problem and leverage gradient-based meta-learning algorithms to solve this task.
We demonstrate that this approach performs on par with auto-decoder based approaches while being an order of magnitude faster at test-time inference.
We further demonstrate that the proposed gradient-based method outperforms encoder-decoder based methods that leverage pooling-based set encoders.
\end{abstract}

\newcommand{\task}{\mathcal{T}}
\newcommand{\VS}[1]{\textbf{\textcolor{red}{[VS: #1]}}}

\newcommand*\Let[2]{\State #1 $\gets$ #2}
\newcommand{\hyponet}[2]{\Phi(#1;#2)}
\newcommand*{\Z}{\makebox[2ex]{\textbf{$\cdot$}}}%
\algrenewcommand\algorithmicrequire{\textbf{Precondition:}}
\algrenewcommand\algorithmicensure{\textbf{Postcondition:}}

\newcommand{\metatrainset}{X_i^{\textit{train}}}
\newcommand{\metatestset}{X_i^{\textit{test}}}

\newcommand{\metatestloss}{\mathcal{L}_{\textit{test}}}
\newcommand{\metatrainloss}{\mathcal{L}_{\textit{train}}}

\section{Introduction}
Humans possess an impressive intuition for 3D shapes; given partial observations of an object we can easily imagine the shape of the complete object.
Computer vision and machine learning researchers have long sought to reproduce this ability with algorithms.
An emerging class of neural implicit shape representations, for example using signed distance functions parameterized by neural networks, promises to achieve these abilities by learning priors over neural implicit shape spaces~\cite{park2019deepsdf,mescheder2019occupancy}.
In such methods, each shape is represented by a function $\Phi$, e.g., a signed distance function. Generalizing across a set of shapes thus amounts to learning a prior over the space of these functions $\Phi$. Two questions arise: (1) How do we parameterize functions $\Phi$, and (2) how do we infer the parameters of such a $\Phi$ given a set of (partial) observations?

Existing methods assume that the space of functions $\Phi$ is low-dimensional and represent each shape as a latent code, which parameterizes the full function $\Phi$ via concatenation-based conditioning or hypernetworks.
These latent codes are either directly inferred by a 2D or 3D convolutional encoder, taking either a single image or a classic volumetric representation as input, or via the auto-decoder framework, where separate latent codes per training sample are treated as free variables at training time. 
Convolutional encoders, while fast, require observations on a regular grid, and are not equivariant to 3D transformations~\cite{finzi2020generalizing}. They further do not offer a straightforward way to accumulate information from a variable number of observations, usually resorting to permutation-invariant pooling of per-observation latent codes~\cite{eslami2018neural}.
Recently proposed pooling-based set encoders may encode sets of variable cardinality~\cite{qi2017pointnet, garnelo2018conditional}, but have been found to underfit the context.
This is corroborated by theoretical evidence, showing that to guarantee universality of the embedded function, the embedding requires a dimensionality of at least the number of context points~\cite{wagstaff2019limitations}.
In practice, most approaches thus leverage the auto-decoder framework for generalization, which is agnostic to the number of observations and does not require observations on a regular grid. This has yielded impressive results on few-shot reconstruction of geometry, appearance and semantic properties~\cite{sitzmann2019srns,kohli2020inferring}.

To infer the parameters of a single $\Phi$ from a set of observations at test time, encoder-based methods only require a forward pass. In contrast, the auto-decoder framework does not learn to infer an embedding from observations, and instead requires solving an optimization problem to find a low-dimensional latent embedding at test time, which may take several seconds even for simple scenes, such as single 3D objects from the ShapeNet dataset.

In this work, we identify a key connection between learning of neural implicit function spaces and meta-learning. 
We then propose to leverage recently proposed gradient-based meta-learning algorithms for the learning of shape spaces, with benefits over both encoder and auto-decoder based approaches.
Specifically, the proposed approach performs on par with auto-decoder based models while being an order of magnitude faster at inference time, does not require observations on a regular grid, naturally interfaces with a variable number of observations, outperforms pooling-based set-encoder approaches, and does not require the assumption of a low-dimensional latent space.

\section{Related Work}
\paragraph{Neural Implicit Scene Representations.} 
Implicit parameterizations of scenes are an emerging topic of interest in the machine learning community.
Parameterized as multilayer perceptrons, these continuous representations have applications in modeling shape parts~\cite{genova2019learning,genova2019deep}, objects~\cite{park2019deepsdf,michalkiewicz2019implicit,atzmon2019sal,gropp2020implicit}, or scenes~\cite{sitzmann2019srns,jiang2020local,peng2020convolutional}. 
These and related neural implicits are typically trained from 3D data~\cite{park2019deepsdf,mescheder2019occupancy,chen2019learning,michalkiewicz2019implicit,atzmon2019sal,gropp2020implicit,jiang2020local,peng2020convolutional,saito2019pifu}, but more recent work has also shown how 2D image data can be directly used to supervise the training procedure~\cite{sitzmann2019srns,Oechsle2019ICCV,Niemeyer2020CVPR,mildenhall2020nerf}, leveraging differentiable neural rendering~\cite{tewari2020state}. 
Earlier work on compositional pattern--producing networks explored similar strategies to parameterize 2D images~\cite{stanley2007compositional,mordvintsev2018differentiable}.

\paragraph{Learning shape spaces.}
A large body of prior work has explored learning priors over the parameters of classic geometry representations, such as meshes, voxel grids, or point clouds~\cite{Maturana2015,Riegler2017OctNet,qi2017pointnet,Jack2018}.
Learning a prior over neural implicit representations of geometry, however, requires learning a prior over a space of \emph{functions}.
To this end, existing work assumes a low-dimensional latent shape space, and leverages auto-decoders or convolutional encoders to regress a latent embedding. This embedding is then decoded into a function either via concatenation-based conditioning~\cite{mescheder2019occupancy,park2019deepsdf} or via hypernetworks~\cite{ha2016hypernetworks, sitzmann2019srns}.
In this work, we instead propose a meta-learning-based approach to learning a shape of signed distance functions.

\paragraph{Meta-Learning.} 
Meta-learning is usually formalized in the context of few-shot learning. Here, the goal is to train a learner that can quickly adapt to new, unseen tasks given only few training examples, often referred to as context observations.
One class of meta-learners proposes to learn an optimizer or update rule~\cite{schmidhuber1987evolutionary,andrychowicz2016learning, ravi2016optimization}. 
Conditional and attentive neural processes~\cite{garnelo2018conditional,eslami2018neural,kim2019attentive} instead encode context observations into a low-dimensional embedding via a permutation-invariant set encoder, and a decoder network is conditioned on the resulting latent embedding. 
A recent class of algorithms proposes to learn the initialization of a neural network, which is then specialized to a new task via few steps of gradient descent~\cite{finn2017model,nichol2018reptile}.
This class of algorithms has recently seen considerable interest, resulting in a wide variety of extensions and improvements. 
\citet{rusu2018meta} blend feed-forward and gradient-descent based specialization via optimization in a latent space. \citet{rajeswaran2019meta} propose to obtain gradients via an implicit method instead of backpropagating through unrolled iterations of the inner loop, leading to significant memory savings.
In this work, we leverage this class of gradient-descent based meta learners to tackle the learning of a shape space.
We refer the reader to a recent survey paper for an exhaustive overview~\cite{hospedales2020meta}.

\section{Meta-learning Signed Distance Functions}
We are interested in learning a prior over implicit shape representations.
As in~\cite{park2019deepsdf}, we consider a dataset $\mathcal{D}$ of $N$ shapes. Each shape is itself represented by a set of points $X_i$ consisting of $K$ point samples from its ground-truth signed distance function (SDF) $\textsc{sdf}_i$:
\begin{align}
    \mathcal{D} = \{X_i\}_{i=1}^N, \quad X_i = \{(\mathbf{x}_j, s_j): s_j = \textsc{sdf}_i(\mathbf{x}_j)\}_{j=1}^K
\end{align}
Where $\mathbf{x}_j$ are spatial coordinates, and $s_j$ are the signed distances at these spatial coordinates.
We aim to represent a shape by directly approximating its SDF with a neural network $\Phi_i$~\cite{park2019deepsdf}, which represents the surface of the shape implicitly as its zero-level set $L_0(\Phi_i)$:
\begin{align}
    L_0(\Phi_i) = \{ x \in \mathbb{R}^3  | \Phi_i(x) = 0 \}, \quad \Phi_i: \mathbb{R}^3 \mapsto \mathbb{R}
\end{align}
We may alternatively choose to approximate the binary `occupancy' of points rather than their signed distances; the surface would then be represented by the binary classifier's decision boundary~\cite{mescheder2019occupancy}.

Generalizing across a set of shapes thus amounts to learning a prior over the space of functions $\mathcal{X}$ where $\Phi_i \in \mathcal{X}$.
Two questions arise: (1) How do we parameterize a single element $\Phi_i$ of the set $\mathcal{X}$, and (2) how do we infer the parameters of such a $\Phi_i$ given a set of (partial) observations $X_i$?
\paragraph{Parameterizing $\Phi_i$: Conditioning via concatenation and hypernetworks.}

Existing approaches to generalizing over shape spaces rely on decoding latent embeddings of shapes. 
In conditioning via concatenation, the target coordinates $\mathbf{x}$ are concatenated with the latent shape embedding $\mathbf{z}_i$ and are fed to a feedforward neural network whose parameters are shared across all shape instances. The latent code conditions the output of the shared network and allows a single shared network to represent multiple shape instances~\cite{mescheder2019occupancy,park2019deepsdf}.

In an alternative formulation, we can instead use a hypernetwork, which takes as input the latent code $\mathbf{z}_i$ and generates the parameters of a shape-representing MLP, $\Phi_i$. 
$\Phi_i$ can then be sampled at coordinates $\mathbf{x}$ to produce output signed distances. 
As we demonstrate in the supplement, conditioning via concatenation is a special case of a hypernetwork~\cite{ha2016hypernetworks}, where the hypernetwork is parameterized as a single linear layer that predicts only the biases of the network $\Phi_i$. Indeed, recent related work has demonstrated that hypernetworks mapping a latent code $\mathbf{z}_i$ to \emph{all} parameters of the MLP $\Phi_i$ can similarly be used to learn a space of shapes~\cite{sitzmann2019srns}.

\paragraph{Inferring parameters of $\Phi_i$ from observations: Encoders and Auto-decoders.}

Existing approaches to inferring latent shape embeddings $\mathbf{z}_i$ rely on encoders or auto-decoders. 
The former rely on an encoder, such as a convolutional or set encoder, to generate a latent embedding that is then decoded into an implicit function using conditioning via concatenation or hypernetworks.

In auto-decoder (i.e., decoder-only) architectures, the latent embeddings are instead treated as learned parameters rather than inferred from observations at training time. At training time, the auto-decoder framework enables straightforward generalization across a set of shapes.
At test time, we freeze the weights of the model and perform a search to find a latent code $\hat{\mathbf{z}}$ that is compliant with a set of context observations $\hat{X}$. A major weakness of the auto-decoder framework is that solving this optimization problem is slow, taking several seconds per object. 

\subsection{Shape generalization as meta-learning}
\label{subsec:shape_as_metalearning}

Meta-learning aims to learn a model that can be quickly adapted to new tasks, possibly with limited training data.
When adapting to a new task, the model is given `context' observations, with which it can modify itself, e.g. through gradient descent. 
The adapted model can then be used to make predictions on unseen `target' observations. 
Formally, supervised meta-learning assumes a distribution over tasks $\task \sim p(\task)$, where each task is of the form $\mathcal{T}=(\mathcal{L}, \{(\mathbf{x}_l, \mathbf{y}_l)\}_{l \in \mathbf{C}}, \{ (\mathbf{x}_j, \mathbf{y}_j)\}_{j \in \mathbf{T}} )$, with loss function $\mathcal{L}$, model inputs $\mathbf{x}$, model outputs $\mathbf{y}$,  the set of all indices belonging to a context set $\mathbf{C}$, and the set of all indices belonging to a target set $\mathbf{T}$. At training time, a batch of tasks $\task_i$ is drawn, with each task split into `context' and `target' observations.
The model adapts itself using the `context' observations and makes predictions on the `target' observations. The model parameters are optimized to minimize the losses $\mathcal{L}$ on all target observations in the training set, over all tasks.

Our key idea is to view the learning of a shape space as a meta-learning problem. In this framework, a task $\mathcal{T}_i$ represents the problem of finding the signed distance function $\Phi_i$ of a shape. Simply put, by giving a model a limited number of `context' observations, each of which consists of a world coordinate location $\mathbf{x}$ and a ground truth signed-distance $s$, we aim to quickly specialize it to approximating the underlying $\textsc{sdf}_i$.
Context and target observations are samples from an object-specific dataset $X_i$: $\mathcal{T}_i=(\{(\mathbf{x}_l, s_l)\}_{l \in \mathbf{C}}, \{ (\mathbf{x}_j, s_j)\}_{j \in \mathbf{T}} )$.
Because each task consists of fitting a signed distance function, we can choose a global loss, such as $\ell_1$.

Note that the auto-decoder framework could be viewed as a meta-learning algorithm as well, though it is not generally discussed as such. 
In this view, the auto-decoder framework is an outlier, as it does not perform model specialization in the forward pass. Instead, specialization requires stochastic gradient descent until convergence, both at training and at test time.
\begin{center}
        \begin{algorithm}[tb]
        \caption{MetaSDF: Gradient-based meta-learning of shape spaces \label{alg:meta_sgd_for_sdfs}}
        \centering
        \begin{algorithmic}[1]
            \small
            \Require{Distribution $\mathcal{D}$ over \textsc{sdf} samples, outer learning rate $\beta$, number of inner-loop steps $k$}
            \State{Initialize inner, per-parameter learning rates $\alpha$, meta-parameters $\theta$} 
                \While{not done}
                \State{Sample batch of shape datasets $X_i\sim \mathcal{D}$}
                \For{\textbf{all} $X_i$}
                    \State{Split $X_i$ into $\metatrainset$, $\metatestset$}
                    \State{Initialize $\phi_i^0=\theta$, $\metatrainloss=0$}
                        \For{$j=0$ \textbf{to} $k$}
                            \Let{$\metatrainloss$}{$\frac{1}{|\metatrainset|} \sum\limits_{(\mathbf{x}, s) \in \metatrainset} \ell_1\left(\hyponet{\mathbf{x}}{\phi_i^j}, s\right)$}
                            \Let{$\phi_i^{j+1}$}{$\phi_i^j - \alpha \odot \nabla_{\phi_i^j} \metatrainloss$}
                         \EndFor
                             \Let{$\metatestloss$}{$\metatestloss + \frac{1}{|\metatestset|} \sum\limits_{(\mathbf{x}, s) \in \metatestset} \ell_1\left(\hyponet{\mathbf{x}}{\phi_i^{k}}, s\right)$}
                \EndFor
                \Let{$\theta, \alpha$}{$(\theta, \alpha) - \beta \nabla_{(\theta, \alpha)} \metatestloss$}
                \EndWhile
                \Return{$\theta$, $\alpha$}
                \end{algorithmic}
        \end{algorithm}
       \vspace{-10pt}
\end{center}

\paragraph{Learning a shape space with gradient-based meta-learning.}
We propose to leverage the family of MAML-like algorithms~\cite{finn2017model}.
We consequently view an instance-specific $\textsc{sdf}_i$, approximated by $\Phi_i$, as the specialization of an underlying meta-network with parameters $\theta$. 
In the forward pass, we sample a batch of shape datasets $X_i$, and split each dataset into a training set $X_i^\textit{train}$ and test set $X_i^\textit{test}$. 
We then perform $k$ gradient descent steps according to the following update rule:
\begin{align}
    \phi_i^{j+1} = \phi_i^j - \lambda \nabla \sum_{(\mathbf{x}, s) \in \metatrainset} \mathcal{L}(\hyponet{\mathbf{x}}{\phi_i^j},s), \quad \phi_i^0 = \theta    
\end{align}
where $\phi_i^j$ are the parameters for shape $i$ at inner-loop step $j$, and $\hyponet{\Z}{\phi_i^j}$ indicates that the network $\Phi$ is evaluated with these parameters.
The final specialized parameters $\phi_i^{k}$ are now used to make predictions on the test set, $\metatestset$, and to compute an outer-loop loss $\metatestloss$.
Finally, we backpropagate the loss $\metatestloss$ through the inner-loop update steps to the parameters $\theta$. 
We further found that the proposed method benefited from the added flexibility of per-parameter learning rates, as proposed by~\citet{li2017meta}.
The full algorithm is formalized in \Cref{alg:meta_sgd_for_sdfs}.

This formulation has several advantages over the auto-decoder framework.
First, as we demonstrate in~\Cref{sec:analysis}, inference at test time is an order of magnitude faster, while performing on par or slightly better both qualitatively and quantitatively.
Further, as MAML optimizes the inference algorithm as well, it can be trained to infer the specialized network from different kinds of context observations. We will demonstrate that this enables, for instance, reconstruction of an SDF given only points on the zero-level set, while the auto-decoder framework requires heuristics in the form of surrogate losses to achieve this goal.
We empirically show that we outperform pooling-based set-encoder based methods, consistent with recent work that found these encoders to underfit the context~\cite{kim2019attentive}.
Finally, both auto-decoder and auto-encoder-based methods assume a low-dimensional latent space, while MAML naturally optimizes in the high-dimensional space of all parameters $\theta$ of the meta-network.

\section{Analysis}
\label{sec:analysis}
In this section, we first apply the proposed MetaSDF approach to the learning of 2D SDFs extracted from MNIST digits, and subsequently, on 3D shapes from the ShapeNet dataset.
All code and datasets will be made publicly available.
\begin{wrapfigure}[7]{R}{0.25\textwidth}
        \vspace{-0.9cm}
        \includegraphics[width=0.25\textwidth]{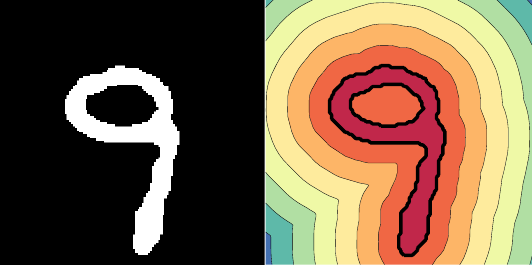}
        \caption{MNIST digit with extracted SDF, bold zero-level set.}
        \label{fig:mnist_2d_sdf}
        \vspace{-.5cm}
\end{wrapfigure}
\subsection{Meta-learning 2D Signed Distance Functions}
\label{subsec:2D_SDF}
We study properties of different generalization methods on 2D signed distance functions (SDFs) extracted from the MNIST dataset.
From every MNIST digit, we extract a 2D SDF via a distance transform, such that the contour of the digit is the zero-level set of the corresponding SDF, see Fig.~\ref{fig:mnist_2d_sdf}.
Following~\cite{park2019deepsdf}, we directly fit the SDF of the MNIST digit via a fully connected neural network.
We benchmark three alternative generalization approaches. 
First, two auto-decoder based approaches, where the latent code is decoded into a function $\Phi$ either using conditioning via concatenation as in~\cite{park2019deepsdf} or via a fully connected hypernetwork as in~\cite{sitzmann2019srns}.
Second, we compare to a conditional neural process (CNP)~\cite{garnelo2018conditional}, representative of permutation-invariant set-encoders.

\begin{figure*}[t!]
        \centering
        \begin{minipage}[t]{0.48\textwidth}
                \vspace{0pt}
                \centering
                \includegraphics[width=\textwidth]{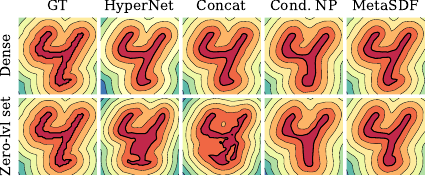}
        \end{minipage} \hfill           
        \begin{minipage}[t][][b]{0.47\textwidth}
                \vspace{10pt}
                \centering
                \begin{tabularx}{\textwidth}{@{}lcccc@{}}
                        \toprule
                        & \phantom{} & Dense & \phantom{} & Zero-level set \\
                        \midrule
                        HyperNet && \num{0.044} && \num{1.562} \\
                        Concat && \num{0.050} && \num{1.813} \\
                        Cond. NP && \num{0.108} && \num{0.156} \\
                        MetaSDF && \textbf{0.015} && \textbf{0.057} \\
                        \bottomrule
                \end{tabularx}
                \label{tbl:context_quant}
        \end{minipage}
        \caption{Qualitative and quantitative comparison of performance of different generalization strategies in inferring the signed distance function of an MNIST digit either from dense observations (top row), or only points on the zero-level set (bottom row). All quantitative results are $\ell_1$-error  $\times 10^{-3}$.}
        \label{fig:context_qual}
        \vspace{-0.25cm}
\end{figure*}

\paragraph{Implementation.}
All models are implemented as fully connected ReLU-MLPs with $256$ hidden units and no normalization layers.
$\Phi$ is implemented with four layers. The set encoder of CNPs similarly uses four layers. Hypernetworks are implemented with three layers as in~\cite{sitzmann2019srns}.
The proposed approach performs $5$ inner-loop update steps, where we initialize $\alpha$ as \num{1e-1}.
All models are optimized using the ADAM optimizer~\cite{kingma2014adam} with a learning rate of \num{1e-4}.

\paragraph{Inference with partial observations.}
We demonstrate that we may train the MetaSDF to infer continuous SDFs either from dense ground-truth samples, or from samples from \emph{only the zero-level set}.
This is noteworthy, as inferring an SDF from only the zero-level set points amounts to solving a boundary value problem defined by a particular \emph{Eikonal equation}, see~\cite{gropp2020implicit}.
In DeepSDF~\cite{park2019deepsdf}, the authors also demonstrate test-time reconstruction from zero-level set points, but require augmenting the loss with heuristics to ensure the problem is well posed. Gropp et al.~\cite{gropp2020implicit} instead propose to explicitly account for the Eikonal constraint in the loss.
We train all models on SDFs of the full MNIST training set, providing supervision via a regular grid of $64\times64$ ground-truth SDF samples. 
For CNPs and the proposed approach, we train two models each, conditioned on either (1) the same $64\times64$ ground-truth SDF samples or (2) a set of $512$ points sampled from the zero-level set.
We then test all models to reconstruct SDFs from the unseen MNIST test set from these two different kinds of context.
Fig.~\ref{fig:context_qual} shows a qualitative and quantitative comparison of results.
All approaches qualitatively succeed in reconstructing SDFs with a dense context set. Quantitatively, MetaSDFs perform on par with auto-decoder based methods for conditioning on dense SDF values. Both methods outperform CNPs by an order of magnitude. This is consistent with prior work showing that pooling-based set encoders tend to underfit the context set. 
When conditioned on the zero-level set only, auto-decoder based methods fail to reconstruct test SDFs. In contrast, both CNPs and MetaSDFs succeed in reconstructing the SDF, with the proposed approach again significantly outperforming CNPs.

\paragraph{Inference speed.}
We compare the time required to fit a single unseen 2D MNIST SDF for both hypernetworks and the concatenation-based approach. 
Even for this simple example, both methods require on the order of $4$ seconds to converge on the latent code of an unseen SDF at test time. 
In contrast, the proposed meta-learning based approach infers functional representations in 5 gradient descent steps, or in about $50$~ms.

\begin{figure*}[t]
        \centering
        \begin{minipage}[t]{0.48\textwidth}
                \centering
                \includegraphics[width=\textwidth]{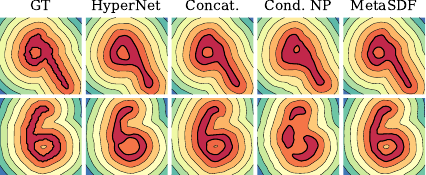}
                \caption{Reconstruction of SDFs of two test digits from classes unseen at training time.}
                \label{fig:qual_unseen}
        \end{minipage} \hfill           
        \begin{minipage}[t]{0.48\textwidth}
                \centering
                \includegraphics[width=\textwidth]{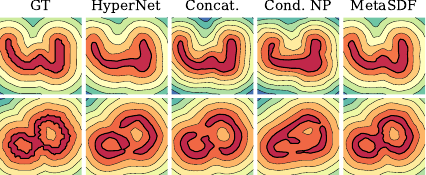}
                \caption{Reconstruction of SDFs of two rotated test digits.}
                \label{fig:qual_rotation}
        \end{minipage} \hfill           
        \begin{minipage}[b]{0.48\textwidth}
                \centering
                \includegraphics[width=\textwidth]{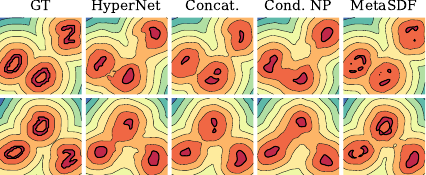}
                \caption{Reconstruction of SDFs of two examples with three digits.}
                \label{fig:qual_compositionality}
        \end{minipage} \hfill           
        \begin{minipage}[b]{0.47\textwidth}
                \centering
                \small
                \setlength\tabcolsep{2.8pt}
                \begin{tabularx}{\textwidth}{@{}lcccccc@{}}
                        \toprule
                        & \phantom{} & Unseen Digit & \phantom{} & Rotation & \phantom{} & Composit. \\
                        \midrule
                        HyperNet && \num{0.144} && \num{0.189} && \num{0.284} \\
                        Concat && \num{0.062} && \num{0.151} && \num{0.137} \\
                        Cond. NP && \num{0.253} && \num{0.316} && \num{0.504} \\
                        MetaSDF && \textbf{0.023} && \textbf{0.032} && \textbf{0.059} \\
                        \bottomrule
                \end{tabularx}
                \captionof{table}{$\ell_1$-error for reconstruction of out-of-distribution samples. All results are $\times 10^{-3}$.}
                \label{tbl:quant_out_of_distribution}
        \end{minipage}
\end{figure*}

\paragraph{Out-of-distribution generalization.}
We investigate the capability of different optimization methods to reconstruct out-of-distribution test samples.
We consider three modes of out-of-distribution generalization: First, generalization to MNIST digits unobserved at training time. Here, we train on digits 0--5, holding out 6--9. 
Second, generalization to randomly rotated MNIST SDFs, where we train on the full MNIST training set. 
And last, generalization to signed distance functions extracted from the triple-MNIST dataset~\cite{mulitdigitmnist}, a dataset of compositions of three scaled-down digits into a single image. 
In this last experiment, we train on a dataset comprised of both the full MNIST training set, as well as the double-MNIST dataset, testing on the triple-MNIST dataset. The double-MNIST dataset contains digits of the same size as the target triple-MNIST dataset.
Figures~\ref{fig:qual_unseen},~\ref{fig:qual_rotation}, and~\ref{fig:qual_compositionality} show qualitative results, while Table~\ref{tbl:quant_out_of_distribution} reports quantitative performance.
The proposed MetaSDF approach far outperforms all alternative approaches in this task. Qualitatively, it succeeds in generating both unseen digits and rotated digits, although no method succeeds to accurately reconstruct the zero-level set in this compositionality experiment.
\begin{wrapfigure}[10]{R}{0.25\textwidth}
  \vspace{-0.6cm}
  \includegraphics[width=0.25\textwidth]{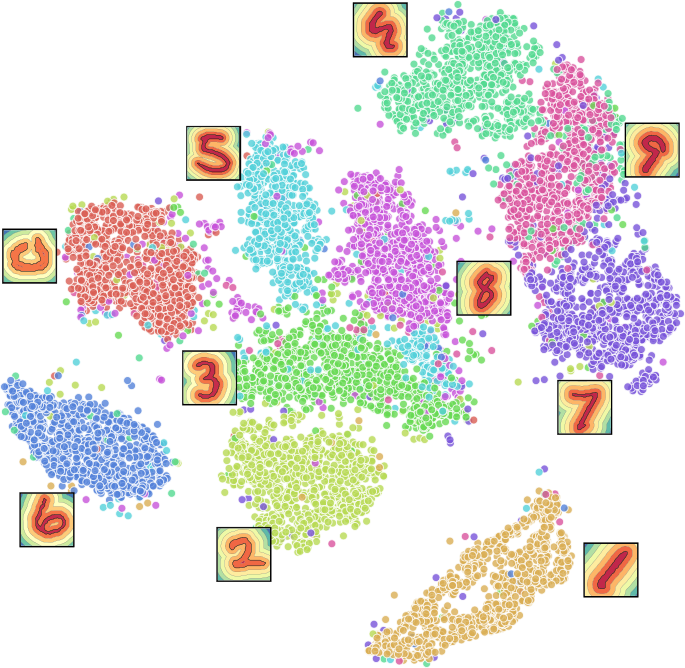}
  \caption{T-SNE embedding of \emph{parameters} of specialized  networks.}
  \label{fig:tsne_mnist}
\end{wrapfigure}
This suggests that the proposed approach is more flexible to out-of-distribution samples.

\paragraph{Interpretation as representation learning.} 
It has previously been observed that learning implicit representations can be seen as a form of representation learning~\cite{kohli2020inferring}. 
Here, we corroborate this observation, and demonstrate that weights of specialized SDF networks found with the proposed meta-learning approach encode information about the digit class, performing unsupervised classification.
Fig.~\ref{fig:tsne_mnist} shows T-SNE~\cite{maaten2008visualizing} embeddings of the parameters of $10,000$ test-set MNIST SDFs. 
While classes are not perfectly linearly separable, it is apparent that the parameters of the SDF-encoding neural network $\Phi$ carry information about the class of the MNIST digit.

\subsection{Meta-Learning a 3D Shape Space}
\begin{figure*}[ht!]
        \centering
    \includegraphics[width=\textwidth]{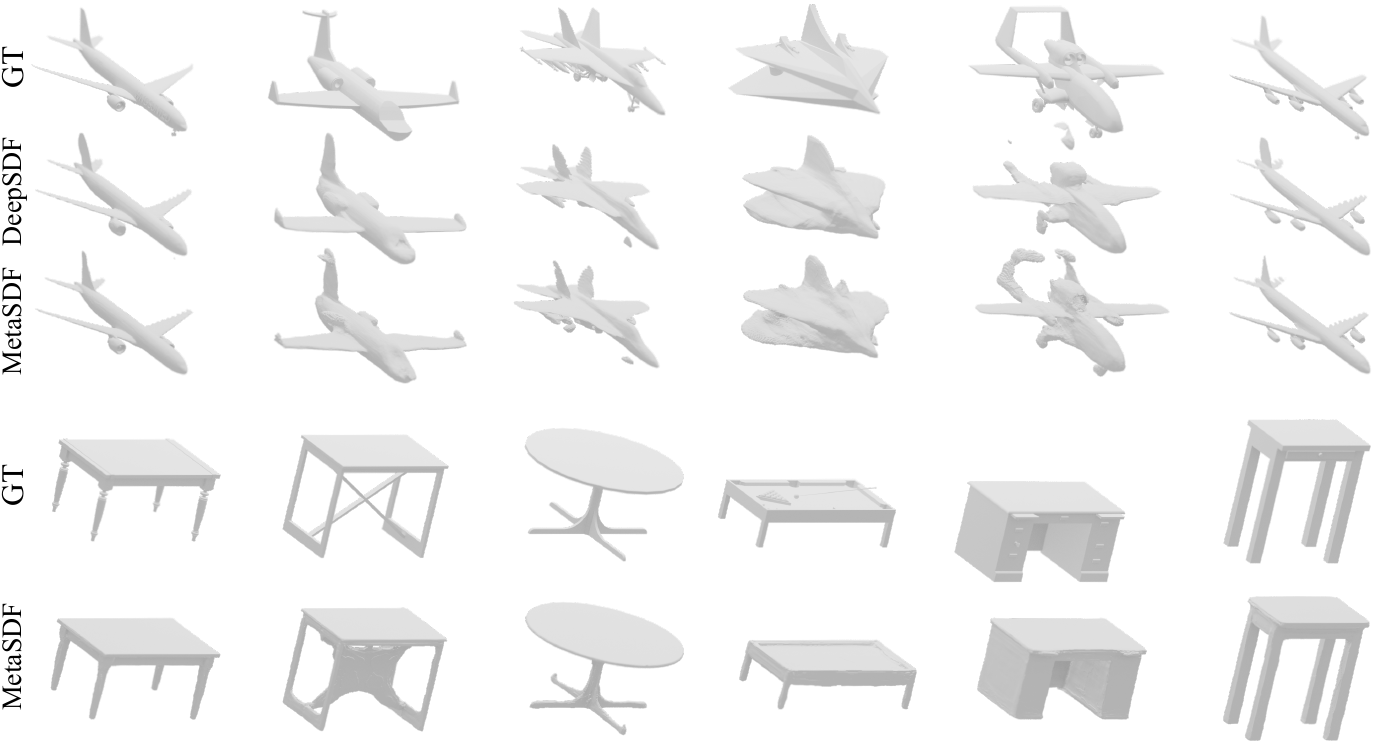}
    \caption{Top row: comparison of reconstructed shapes of DeepSDF and MetaSDF on the ShapeNet planes class. Both models receive dense signed distance values at training and test time. The two methods perform approximately on par, with the proposed method performing slightly better on out-of-distribution samples while speeding up inference by one order of magnitude. Bottom row: Additional reconstructions of MetaSDF from the ShapeNet tables class.}
    \label{fig:deep_sdf_meshes}
\end{figure*}
\begin{figure}
        \centering
        \includegraphics[width=\textwidth]{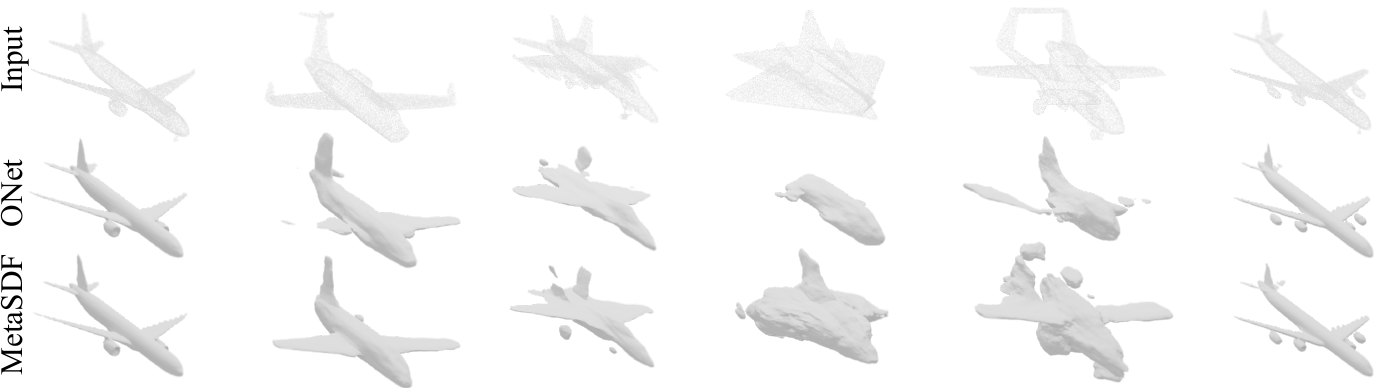}
        \caption{Comparison of reconstructed shapes of proposed method and a PointNet-encoder based method as proposed in~\cite{mescheder2019occupancy}.
        Both models are conditioned only on zero-level set points at training and test time. 
        The proposed method generally outperforms the PointNet-based approach, with strongest gains in uncommon airplanes.}
        \label{fig:onet_comparison}
\end{figure}

\begin{table}
\begin{minipage}[t]{.49\linewidth}
        \centering
        \caption{Mean / median Chamfer Distance (CD) evaluation for ShapeNetV2, conditioned on densely sampled SDFs. All results are $\times 10^{-3}$.}
        \label{tbl:CD_Dense}
        \begin{tabularx}{\linewidth}{@{}lcrc@{}}
                \toprule
                \phantom{}
                & DeepSDF
                & \phantom{}
                & MetaSDF
                \\
                \midrule
                Planes & \num{0.073} / \num{0.031} && \textbf{0.053} / \textbf{0.021}\\
                Tables & \num{0.553} / \num{0.068} && \textbf{0.134} / \textbf{0.059}\\
                \bottomrule
        \end{tabularx}
\end{minipage}
\hfill
\begin{minipage}[t]{.49\linewidth}
        \centering
        \caption{Mean / median Chamfer Distance (CD) evaluation for ShapeNetV2, conditioned on zero-level set points. All results are $\times 10^{-3}$.}
        \label{tbl:CD_Levelset}
        \begin{tabularx}{\linewidth}{@{}lcrc@{}}
                \toprule
                \phantom{}
                & PointNet Enc.
                & \phantom{}
                & MetaSDF
                \\
                \midrule
                Planes & \num{0.558} / \num{0.222} && \textbf{0.139} / \textbf{0.067}\\
                Tables & \num{0.693} / \num{0.216} && \textbf{0.327} / \textbf{0.095}\\
                \bottomrule
        \end{tabularx}
\end{minipage}
\vspace{-0.15cm}
\end{table}

We now demonstrate that the proposed approach scales to three dimensions, where results on 3D shape representation are consistent with the 2D MNIST results above.

We first benchmark the proposed approach in the case of a densely sampled SDF.
To this end, we follow the experimental setup of Park et al.~\cite{park2019deepsdf}, and benchmark with their DeepSDF model.
DeepSDF combines the auto-decoder framework with conditioning via concatenation to learn a prior over 3D shapes in the ShapeNet~\cite{chang2015shapenet} dataset.
The dataset contains ground-truth samples from the analytically computed SDFs of the respective mesh.
The application of the proposed framework to this dataset follows the 2D examples in Sec.~\ref{subsec:2D_SDF}, with two minor differences. 
Park et al.~\cite{park2019deepsdf} propose to clamp the $\ell_1$ loss to increase accuracy of the SDF close to the surface.
We empirically found that clamping in the inner loop of a gradient-based meta-learning algorithm leads to unstable training. 
Instead, we found that a multitask loss function that tracks both SDF values as well as occupancy, i.e. the sign of the signed distance function, provides stable training, increases the level of detail of the reconstructed shapes for all models, and provides faster convergence. 
Instead of predicting a single signed distance output, we predict two values: the distance to the surface and the sign of the distance function (inside or outside the object). 
The $\ell_1$ SDF loss is then combined with a binary cross-entropy sign loss with the uncertainty loss weighting scheme described by Kendall et al. \cite{kendall2018multi}. 
This strategy can be interpreted as combining the loss terms of the concurrently proposed DeepSDF and Occupancy Networks~\cite{park2019deepsdf, mescheder2018occupancy}.
Table~\ref{tbl:CD_Dense} reports mean and median Chamfer distance of reconstructed meshes on the test set, while Fig.~\ref{fig:deep_sdf_meshes} displays corresponding reconstructed shapes.
As in the 2D experiments, the proposed approach generally performs on par with the baseline auto-decoder approach. 
As indicated by the previous out-of-distribution experiments, differences mainly arise in the reconstruction of uncommon shapes (see columns 4 and 5 of Fig.~\ref{fig:deep_sdf_meshes}), where the proposed approach fares better---this is reflected in the gap in mean Chamfer distance, while median Chamfer distance is almost identical.
However, reconstruction of a 3D shape at test time is \emph{significantly} faster, improving by more than one order of magnitude from 8 seconds to 0.4 seconds.
To evaluate the impact of our proposed composite loss, we compare DeepSDF trained using our composite loss function against the same architecture trained using the original $\ell_1$ loss with clamping. 
As opposed to the proposed approach, where the composite loss is critical to stable training, DeepSDF profits little, with mean and median Chamfer Distances on ShapeNet planes at \num{6.7e-5} and \num{3.3e-5}, respectively.

Next, as in our 2D experiments, we demonstrate that the same MetaSDF model---with no changes to architecture or loss function---can learn to accurately reconstruct 3D geometry conditioned only on zero-level set points, i.e., a point cloud.
We compare to a baseline with a PointNet~\cite{qi2017pointnet} encoder and conditioning via concatenation, similar to the model proposed in Occupancy Networks~\cite{mescheder2018occupancy}. We match the parameter counts of both models.
Table~\ref{tbl:CD_Levelset} reports the Chamfer distance of test-set reconstructions for this experiment, while Fig.~\ref{fig:onet_comparison} shows qualitative results.
Consistent with our results on 2D SDFs, the proposed MetaSDF approach outperforms the set-encoder based approach. Again, performance differences are most significant with out-of-distribution shapes.
We note that auto-decoder-based models cannot perform zero-level set reconstructions without the aid of additional heuristics.

\begin{figure*}
        \centering
        \includegraphics[width=\textwidth]{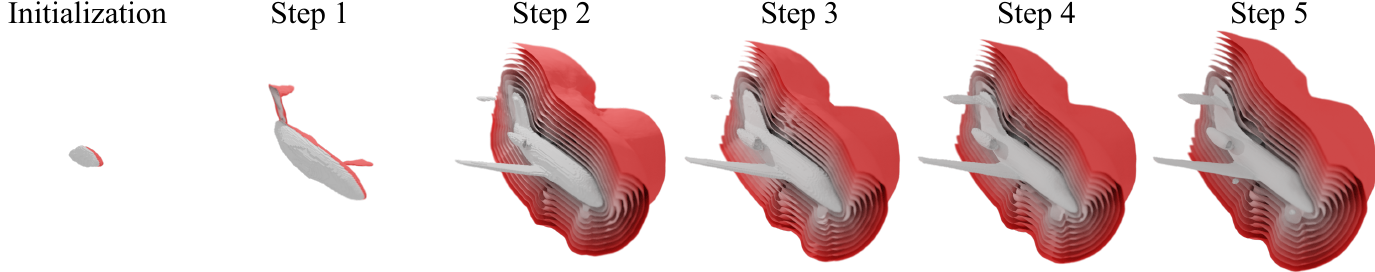}
        \caption{Evaluation of the SDF level sets throughout the inner-loop iterations. Red contours are level sets in increments of 0.05.}
        \label{fig:intermediate}
\end{figure*}

\paragraph{Visualizing inner loop steps.} 
To gain additional insight into the process of the inner-loop optimization, in Fig.~\ref{fig:intermediate}, we visualize the evolution of the signed distance function over inner loop iterations, beginning with the unspecialized meta-network.
The underlying model was trained for level-set inference on the ShapeNet planes class.
Intriguingly, even though this model is class-specific, the initialization does not resemble an airplane, and only encodes valid signed distances starting from the second iteration. 
At initialization and after the first update step, function values increase significantly faster than those of the ground-truth signed distance function, visualized by the collapse of the level sets onto the surface.
In contrast, the mean zero-vector of the auto-decoder based DeepSDF model encodes an object that resembles an ``average plane'' (see supplemental material). Future investigation of this difference may reveal additional insight into how gradient-based meta learning algorithms succeed in conditioning the optimization problem for fast convergence.

\section{Discussion}
In summary, MetaSDF is a meta-learning approach to learning priors over the space of SDFs represented by fully connected neural networks.
This approach performs on par with auto-decoder based approaches while being an order of magnitude faster at inference time, outperforms pooling-based set-encoder methods, and makes weaker assumptions about the dimensionality of the latent space.
However, several limitations remain. The current implementation requires backpropagation through the unrolled inner-loop gradient descent steps, and thus the computation of second-order gradients, which is highly memory-intensive.
This may be addressed by recently proposed implicit gradient methods that offer memory complexity independent of the number of inner-loop optimization steps~\cite{rajeswaran2019meta}.
More generally, significant progress has recently been made in gradient-based meta-learning algorithms. We leverage Meta-SGD~\cite{li2017meta} as the inner-loop optimizer in this paper, but recent work has realized significant performance gains in few-shot learning via optimization in latent spaces~\cite{rusu2018meta} or by learning iterative updates in an infinite-dimensional function space~\cite{xu2019metafun}, which could further improve \mbox{MetaSDFs}. 
Another promising line of future investigation are recently proposed attention-based set encoders~\cite{kim2019attentive}, which have been shown to perform better than set encoders that aggregate latent codes via pooling, though the memory complexity of the attention mechanism makes conditioning on large (5k points) context sets as in our 3D experiments computationally costly.
Finally, future work may apply this approach to representations of scenes, i.e., both 3D geometry and appearance, via neural rendering~\cite{sitzmann2019srns, mildenhall2020nerf}.

Our approach advances the understanding of generalization strategies for emerging neural implicit shape representations by drawing the connection to the meta-learning community. We hope that this approach inspires follow-on work on learning more powerful priors of implicit neural shape representations.

\section*{Broader Impact}
Emerging neural implicit representations are a powerful tool for representing signals, such as 3D shape and appearance.
Generalizing across these neural implicit representations requires efficient approaches to learning distributions over functions.
We have shown that gradient-based meta-learning approaches are one promising avenue to tackling this problem.
As a result, the proposed approach may be part of the backbone of this emerging neural signal representation strategy. 
As a powerful representation of natural signals, such neural implicits may in the future be used for the generation and manipulation of signals, which may pose challenges similar to those posed by generative adversarial models today. 

\begin{ack}
	We would like to offer special thanks to Julien Martel, Matthew Chan, and Trevor Chan for fruitful discussions and assistance in completing this work.
	Vincent Sitzmann was supported by a Stanford Graduate Fellowship.
	Gordon Wetzstein was supported by an NSF CAREER Award (IIS 1553333), a Sloan Fellowship, and a PECASE from the ARO.
\end{ack}

{\small
	\bibliographystyle{unsrtnat}
	\bibliography{metasdf}
}



\section*{Supplementary material}

# 1 Conditioning via concatenation as a special case of Hypernetworks

Here, we demonstrate that conditioning via concatenation is a special case of a hypernetwork, where the hypernetwork is a single affine layer that only predicts the biases of the hyponetwork.

We first formalize a hypernetwork $\Phi$ that predicts the weights of a single layer of some hyponetwork. The hypernetwork maps a code vector $\mathbf{z} \in \mathbb{R}^n$ to the weights $\mathbf{W} \in \mathbb{R}^{m \times l}$ and biases $\mathbf{b} \in \mathbb{R}^l$ of the hypo-layer:

$$\Psi : \mathbb{R}^n \to \mathbb{R}^{mln}, \quad \mathbf{z} \mapsto \Psi(\mathbf{z}) = (\mathbf{W}_{(0,0)}, ..., \mathbf{W}_{(m,l)}, \mathbf{b}_0, ..., \mathbf{b}_m) \quad (1)$$

We now analyze a single layer of a neural network with conditioning via concatenation. This single layer $\mathbf{y}$ has inputs $\mathbf{x} \in \mathbb{R}^k$, $l$ hidden units, a weight matrix $\mathbf{W} \in \mathbb{R}^{l \times (k+n)}$, and bias vector $\mathbf{b} \in \mathbb{R}^l$, and is conditioned on a code vector $\mathbf{z} \in \mathbb{R}^n$ via concatenation with $\mathbf{x}$. We are only interested in the weights and biases and therefore omit the nonlinearity. This can be formalized as follows:

$$\mathbf{y} = \mathbf{W} \cdot (\mathbf{x} \| \mathbf{z}) + \mathbf{b} \quad (2)$$

where $(\cdot \| \cdot)$ signifies concatenation. We can now split up the weight matrix $\mathbf{W}$ by rows into two weight matrices, $\mathbf{W}_{\text{hypo}} \in \mathbb{R}^{l \times k}$ and $\mathbf{W}_{\text{hyper}} \in \mathbb{R}^{l \times n}$, and re-write $\mathbf{y}$ as follows:

$$\mathbf{y} = \mathbf{W}_{\text{hypo}} \cdot \mathbf{x} + \underbrace{\mathbf{W}_{\text{hyper}} \cdot \mathbf{z} + \mathbf{b}}_{\Psi_{\text{bias}}(\mathbf{z})} \quad (3)$$

The affine term $\mathbf{W}_{\text{hyper}} \cdot \mathbf{z} + \mathbf{b}$ does not depend on the input, only on the latent code $\mathbf{z}$, and is additive with the input-dependent linear term $\mathbf{W}_{\text{hypo}} \cdot \mathbf{x}$. We can thus identify it as a "conditional bias" that is computed by a hypernetwork $\Psi_{\text{bias}}(\mathbf{z}) = \mathbf{W}_{\text{hyper}} \cdot \mathbf{z} + \mathbf{b}$, where the parameters $\mathbf{W}_{\text{hyper}}$ and $\mathbf{b}$ are intepreted as parameters of the hypernetwork. Conditioning via concatenation is thus equivalent to a hypernetwork with a single affine layer that only predicts the biases of a hyponetwork.

# 2 2D Experiments

## 2.1 Additional Results

We provide mean and standard deviation for quantitative 2D SDF experiments for the conditional neural process and MetaSDF results in tables 1 and 2. In addition to the conditional neural process with a 4-layer ReLU set encoder, we report mean and standard deviation for a conditional neural process with a 9-layer ReLU set encoder. While the number of layers more than doubled, performance increases only slightly, still lagging far behind the performance of the proposed MetaSDF approach.

Table 1: $\ell_1$-error (mean/std.) for reconstructions from dense and levelset observations. All results are $\times 10^{-6}$.

|  | Dense | Zero-level set |
| --- | --- | --- |
| Cond. NP, 4-layer set encoder | 101.7/5.1 | 154.9/1.0 |
| Cond. NP, 9-layer set encoder | 92.5/2.0 | 145.8/1.2 |
| MetaSDF | **15.4/0.2** | **56.6/0.5** |

## 2.2 Reproducibility

Here, we provide exact specifications of the 2D experiments to ensure reproducibility. All code and datasets will be made publicly available.

Table 2: $\ell_1$-error (mean/std.) for reconstruction of out-of-distribution samples. All results are $\times 10^{-6}$.

|  | Unseen Digit | Rotation | Composit. |
| --- | --- | --- | --- |
| Cond. NP, 4-layer set encoder | 252.4/5.0 | 301.1/10.3 | 484.9/13.6 |
| Cond. NP, 9-layer set encoder | 223.4/4.2 | 291.2/8.6 | 468.0/22.8 |
| MetaSDF | **22.8/0.5** | **33.3/1.1** | **59.8/1.2** |

#### 2.2.1 Datasets

For all single-MNIST experiments, we use the MNIST dataset, as supplied with the pytorch torchvision library (link), as a starting point. We use the official train-test split of the MNIST dataset, and further split 1000 examples off the training set as a validation set. For the double-MNIST and triple-MNIST experiments, we use the datasets provided by [5] (link). The large size of the triple-MNIST test set makes it infeasible to reconstruct it using the auto-decoder framework. We thus only pick 1000 elements of the triple-MNIST dataset as a test set. To extract 2D signed distance functions, we first binarize the greyscale image, and then apply two unsigned 2D distance transforms as implemented in the scipy package, once on the original binary image and once on its inverse binary image, and combine them into a single signed distance function.

#### 2.2.2 Model Details

All models are implemented as fully connected ReLU-MLPs with 256 hidden units and no normalization layers. The SDF network $\Phi$ is implemented with four layers (i.e., two hidden layers). The concatenation-based conditioning is performed by concatenating the latent code once to the input and once to the third layer of $\Phi$. The set encoder of CNPs is implemented with four layers. Hypernetworks are implemented with three layers as in [4], predicting all parameters (weights and biases) of $\Phi$. The proposed approach performs 5 inner-loop update steps, where we initialize $\alpha$ as $1 \times 10^{-1}$.

#### 2.2.3 Out-of-distribution experiments

We conduct all out-of-distribution experiments using models trained on *dense* samples from the signed distance function, and condition on dense samples at test-time as well. We note that baseline approaches performed much better in the dense setting, and this thus enables the fairest comparison of the proposed approach and baseline approaches.

#### 2.2.4 Hyperparameters

We train all models with ADAM with a learning rate of $1 \times 10^{-4}$ and a batch size of 32, sampling $32^2$ points for the dense SDF experiments and 512 points for the levelset experiments. We train encoder- and meta-learning based models for 50 epochs. Due to slower convergence of the auto-decoder framework, we train auto-decoder models for 150 epochs.

#### 2.2.5 Computing Hardware

We train all models on a single NVIDIA RTX 6000 GPU. Hypernetworks, conditioning via concatenation-based models, as well as conditional neural processes consume approximately 3GB of GPU memory at training time. The proposed MetaSDF approach consumes approximately 5GB of GPU memory at training time.

#### 2.2.6 Runtime & Complexity

Conditional neural processes train in about 3 hours, while MetaSDF trains in about 5 hours. The training time of auto-decoder based approaches grows significantly with the size of the dataset. For out-of-distribution experiments such as the rotated MNIST digits or the compositionality experiments, the size of the datasets required 24 hours of training per model, and another 24 hours of test-time reconstruction. The memory complexity of the proposed Meta-Learning approach is $O(nm)$, with the number of query points $n$ and the number of inner gradient descent steps $m$.

3

### 2.2.7 Number of Evaluation Runs

Each method was run on the test set exactly three times to calculate mean and standard deviation of the final result.

## 3 3D Experiments

### 3.1 Additional Results

#### 3.1.1 "Average airplane" of concatenation-based approach vs. Meta-network initialization

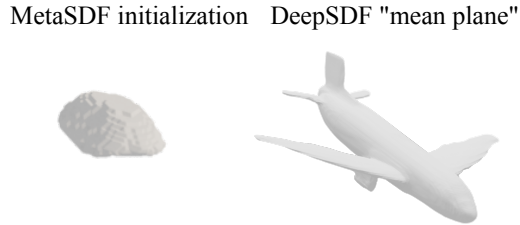

Figure 1: Comparison of the zero-level set of MetaSDF before specialization and the "mean plane" of DeepSDF, i.e., the zero-level set of the airplane encoded by the all-zero latent code.

In Fig. 1, we show the zero-level set of the meta-initialization discovered by the proposed MetaSDF approach as well as the zero-level set of the SDF corresponding to the all-zero "mean latent code" as discovered by DeepSDF [3]. Intriguingly, they bear little resemblance. Future work may investigate the properties of the initialization of MetaSDF, potentially providing insight into why this initialization enables such swift specialization with only five gradient descent steps.

#### 3.1.2 Qualitative comparison to DeepSDF and PointNet encoder on Shapenet V2 Tables

Fig. 2 and Fig. 3 show qualitative comparisons of the proposed MetaSDF approach to DeepSDF and a pointnet-encoder based approach on the Shapenet V2 tables class.

### 3.2 Reproducibility.

Here, we provide exact specifications of the 3D experiments to ensure reproducibility. All code and datasets will be made publicly available.

#### 3.2.1 Datasets.

We utilize the 'Planes' and 'Tables' categories of ShapeNetV2 [1] (link) for all 3D experiments. We preprocess SDF samples, used in dense reconstruction, and surface samples, used in zero-level set reconstruction, with the preprocessing code (link) from DeepSDF [3]. We use DeepSDF's train-test splits for both object categories, but extract 20 examples from each training set for use as validation sets. We encountered errors in using DeepSDF's preprocessing code and thus were only able to process about 80% of the 3D models.

#### 3.2.2 Model Details.

**Details on composite loss** Empirically, we found that a composite loss function that combines the $\ell_1$ loss on the value of the predicted SDF with a binary crossentropy loss on the sign of the predicted SDF lead to better results for MetaSDF. Because meshes are represented by the zero-level set, precisely predicting the sign of the SDF close to the surface is far more important to mesh accuracy than accurately predicting the value of the SDF away from the surface. Intuitively, our strategy encourages the model to emphasize not only the value but also the sign of SDF predictions. Our model predicts two outputs: a prediction of signed distance and a sigmoid output representing the sign. At training time, we calculate the $\ell_1$ loss on the SDF prediction and the binary crossentropy loss on the sign prediction. We combine these loss terms with the method proposed by Kendall et

4

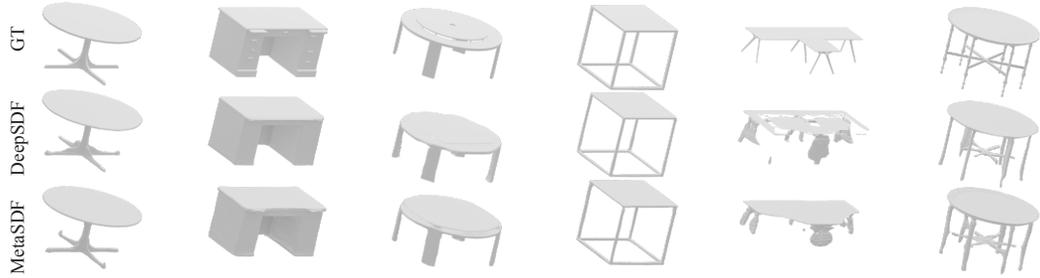

Figure 2: Qualitative comparison of the proposed MetaSDF approach and DeepSDF on the Shapenet v2 tables dataset.

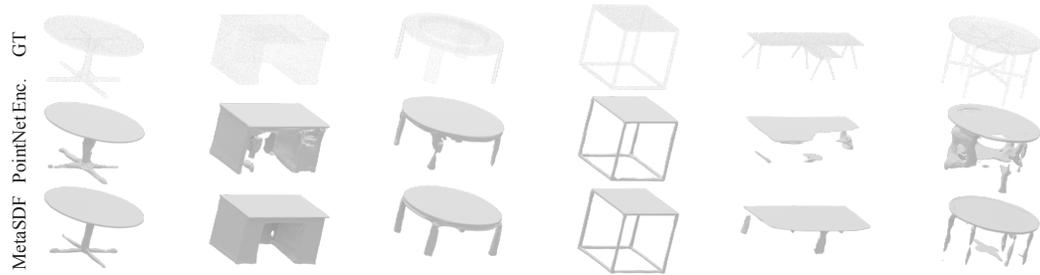

Figure 3: Qualitative comparison of the proposed MetaSDF approach and a PointNet Encoder on zero-level set reconstructions from the Shapenet v2 tables dataset.

al. [2]. At test time, we combine the two outputs by multiplying the absolute value of the distance prediction with the predicted sign to produce a single SDF prediction.

**Dense MetaSDF** MetaSDF for our dense experiments is implemented as a fully connected ReLU-MLP with 8 layers of 512 units and no normalization layers. We utilize 5 inner-loop update steps and per-parameter, per-step learning rates, which are initialized to $5 \times 10^{-3}$.

**Zero-Level set MetaSDF** We utilize a similar architecture for our zero-level set experiments, but in order to better match parameter counts to PointNet encoders, we reduced network size to 7 layers of 512 units.

**DeepSDF** We utilize the official DeepSDF implementation, which consists of 8 layers of 512 units and weight normalization. DeepSDF uses an autodecoder, with latent codes of 256 units.

**PointNet Encoder** We implement a resnet PointNet encoder, which consists of 5 resnet blocks of 512 units and encodes a 256-dimensional latent code. The SDF network, $\Phi$, is implemented with 7 layers of 512 units.

### 3.2.3 Hyperparameters.

We train all models with ADAM and initialize learning rates to $5 \times 10^{-4}$. All models were trained to convergence. We trained Dense MetaSDF for 500 epochs using our composite loss function and decayed learning rate by a factor of 0.5 after 350 steps. All other models, including Zero-level set MetaSDF, were trained using L1 loss for 2000 epochs, decaying learning rate by a factor of 0.5 every 500 steps. All models use a batch size of 64 scenes.

### 3.2.4 Computing Hardware

We train all models on dual NVIDIA RTX 8000 GPUs.

### 3.2.5 Runtime & Complexity.

The complexity of the proposed Meta-Learning approach is $O(nm)$, with the number of query points $n$ and the number of inner gradient descent steps $m$. DeepSDF and PointNet Encoders train to 2000 epochs in approximately 4 days while MetaSDF trains to 2000 epochs in approximately 6 days.

### 3.2.6 Number of Evaluation Runs.

Because we randomly sample context points during mesh reconstruction, reconstructed meshes vary between runs. Following the evaluation procedure of DeepSDF, at test-time each model reconstructs each mesh twice; the better reconstruction is used in computing chamfer distance across the test set. This evaluation procedure was conducted once for each model.



\end{document}